\documentclass{article}
\usepackage{spconf,amsmath,graphicx}

\usepackage{times}
\usepackage{epsfig}
\usepackage{graphicx}
\usepackage{amsmath}
\usepackage{amssymb}
\usepackage{booktabs}

\usepackage{cite}
\usepackage{ctable}
\usepackage{hhline}
\usepackage{booktabs}
\usepackage{subcaption}
\usepackage{csquotes}
\usepackage{multirow}

\usepackage{color, colortbl}
\definecolor{Gray}{gray}{0.9}

\newcolumntype{L}[1]{>{\raggedright\let\newline\\\arraybackslash\hspace{0pt}}m{#1}}
\newcolumntype{C}[1]{>{\centering\let\newline\\\arraybackslash\hspace{0pt}}m{#1}}
\newcolumntype{R}[1]{>{\raggedleft\let\newline\\\arraybackslash\hspace{0pt}}m{#1}}

\usepackage{pifont}
\newcommand{\xmark}{\ding{55}}%

\title{Representation Learning of Vertex Heatmaps for 3D Human Mesh Reconstruction from Multi-view Images}
\name{Sungho Chun$^{\star}$ \qquad Sungbum Park$^{\dagger}$ \qquad Ju Yong Chang$^{\star}$\thanks{This work was supported in part by NCSOFT; and in part by the Institute of Information \& Communications Technology Planning \& Evaluation (IITP) grant funded by the Korea government(MSIT) (No.2021-0-00348, Development of A Cloud-based Video Surveillance System for Unmanned Store Environments using Integrated 2D/3D Video Analysis).}}
\address{$^{\star}$Dept of ECE, Kwangwoon University, Seoul, Korea\\
$^{\dagger}$Netmarble, Seoul, Korea\\
{\small \texttt {\{asw9161,jychang\}@kw.ac.kr},\ \ \texttt {spark0916@netmarble.com}}
}
\begin{document}
%\ninept
%
\maketitle
\begin{abstract}
This study addresses the problem of 3D human mesh reconstruction from multi-view images. Recently, approaches that directly estimate the skinned multi-person linear model (SMPL)-based human mesh vertices based on volumetric heatmap representation from input images have shown good performance. We show that representation learning of vertex heatmaps using an autoencoder helps improve the performance of such approaches. Vertex heatmap autoencoder (VHA) learns the manifold of plausible human meshes in the form of latent codes using AMASS, which is a large-scale motion capture dataset. Body code predictor (BCP) utilizes the learned body prior from VHA for human mesh reconstruction from multi-view images through latent code-based supervision and transfer of pretrained weights. According to experiments on Human3.6M and LightStage datasets, the proposed method outperforms previous methods and achieves state-of-the-art human mesh reconstruction performance.
% The proposed method consists of the vertex heatmap autoencoder (VHA) for representation learning, the body code predictor (BCP) for predicting the latent code of human mesh from multi-view images, and the fitting module for obtaining SMPL parameters.
\end{abstract}
%

%%%%%%%%%%%%%%%%%%%%%
\section{Introduction}
\label{sec:intro}

3D human pose and mesh reconstruction from images is an interesting research topic in image processing and computer vision. It can be used in various applications such as virtual/augmented reality and human motion analysis. To date, various approaches have been proposed to address this task, and they can be roughly divided into deep learning-based~\cite{2018_Kanazawa, 2020_Shin, choi2022learning}, optimization-based~\cite{SMPLify_X_2019_Pavlakos, Bogo_ECCV_2016}, and hybrid approaches~\cite{Chun_2023_WACV, lightcap2021} that combine them.

This paper addresses the task of reconstructing a skinned multi-person linear model (SMPL)-based human mesh~\cite{2015_SMPL} from input multi-view images. To this end, we adopt a hybrid approach, namely, learnable human mesh triangulation (LMT)~\cite{Chun_2023_WACV}. LMT is a two-stage method where volumetric heatmap-based vertex regression and SMPL fitting are performed sequentially. The body prior for realistic human mesh reconstruction is used in the fitting stage but not in the vertex regression stage. Therefore, unseen test poses or lack of training data cause the vertex regression stage to produce incorrect vertex coordinates. As a result, only employing the body prior in the fitting stage has limitations in refining incorrectly estimated vertices, which causes performance degradation. To solve this problem, we learn the body prior for the vertex heatmap and use it for vertex regression. %While the body prior is used for realistic human mesh reconstruction in the fitting stage, the body prior is not used in the vertex regression stage.

The proposed model consists of \emph{Vertex Heatmap Autoencoder (VHA)} for learning the human body prior, \emph{Body Code Predictor (BCP)} producing the latent code for a human mesh from multi-view images, and the fitting module that predicts the SMPL parameters. For representation learning~\cite{ radford2015unsupervised, goodfellow2016deep}, we train an autoencoder that uses vertex heatmaps as its input and output. The weights of the trained autoencoder and the latent code computed from the autoencoder implicitly contain prior knowledge on the plausible human body. BCP is trained to predict the latent code computed from VHA. We also initialize the weights of BCP with the weights of the pretrained VHA encoder. BCP can utilize the body prior learned by VHA for vertex regression through the above two processes. We feed the latent code computed from BCP into the VHA decoder to obtain human mesh vertices. The fitting module predicts SMPL parameters from mesh vertices generated by the BCP and VHA decoder. %and apply the loss to the resultant vertices. The use of this additional vertex loss improves the performance of vertex regression

The human body prior has been used in many existing studies for SMPL-based human mesh reconstruction from images, but their prior is different from ours. Deep learning-based methods~\cite{2018_Kanazawa, choi2022learning} use either adversarial training or the pretrained variational autoencoder (i.e., VPoser~\cite{SMPLify_X_2019_Pavlakos}). In optimization-based methods~\cite{Chun_2023_WACV, lightcap2021, SMPLify_X_2019_Pavlakos, Bogo_ECCV_2016}, prior knowledge is generally utilized using VPoser and manually designed pose regularizing terms. These methods allow a plausible human body to be reconstructed by constraining the SMPL parameters. By contrast, our method constrains the human mesh vertices, specifically their volumetric heatmaps.

The multi-person pose estimation method proposed in~\cite{fabbri2020compressed} is similar to ours in that it performs representation learning through an autoencoder. However, its motivation is to solve the problem of excessive computation and memory usage due to volumetric heatmaps, which is different from the motivation of our method. It is also based on representation learning for skeletal joints rather than body mesh vertices, which is another noticeable difference from our method.

The contributions of our study are as follows. First, an autoencoder-based representation learning method for vertex heatmaps is proposed for the first time to our knowledge. Second, large-scale motion capture datasets such as AMASS~\cite{moshpp_AMASS_2019} can be used for representation learning, which helps improve the generalization performance of the proposed mesh reconstruction method. Third, our mesh reconstruction model can be efficiently trained with limited data compared with existing methods. This training data efficiency of our model is due to the proposed body prior. Finally, we show the effectiveness of the proposed method by achieving state-of-the-art human mesh reconstruction performance on Human3.6M~\cite{2014_H36M} and LightStage~\cite{peng2021neural} datasets.
%Third, the training data efficiency of our mesh reconstruction model is relatively high thanks to the proposed body prior. Our model can be efficiently trained with limited data compared to existing methods.
% ~\cite{2018_Kanazawa, 2019_Kolotouros_CVPR, 2019_Kolotouros_ICCV, 2020_I2L, 2021_Lin_CVPR, 2021_Lin_ICCV, cho2022FastMETRO, Choi_2020_ECCV_Pose2Mesh, 2020_Shin, choi2022learning, choi2020beyond, kolotouros2021prohmr, kocabas2019vibe, wang2021mvp, Kocabas_PARE_2021, pymaf2021}
% ~\cite{2018_Kanazawa, 2020_Shin, choi2022learning}

%%%%%%%%%%%%%%%%%%%%%
\section{Proposed Method}
\label{sec:proposed_method}

\subsection{Overview}
\label{ssec:overview}

\begin{figure}[t]
\centering
\includegraphics[width=\linewidth]{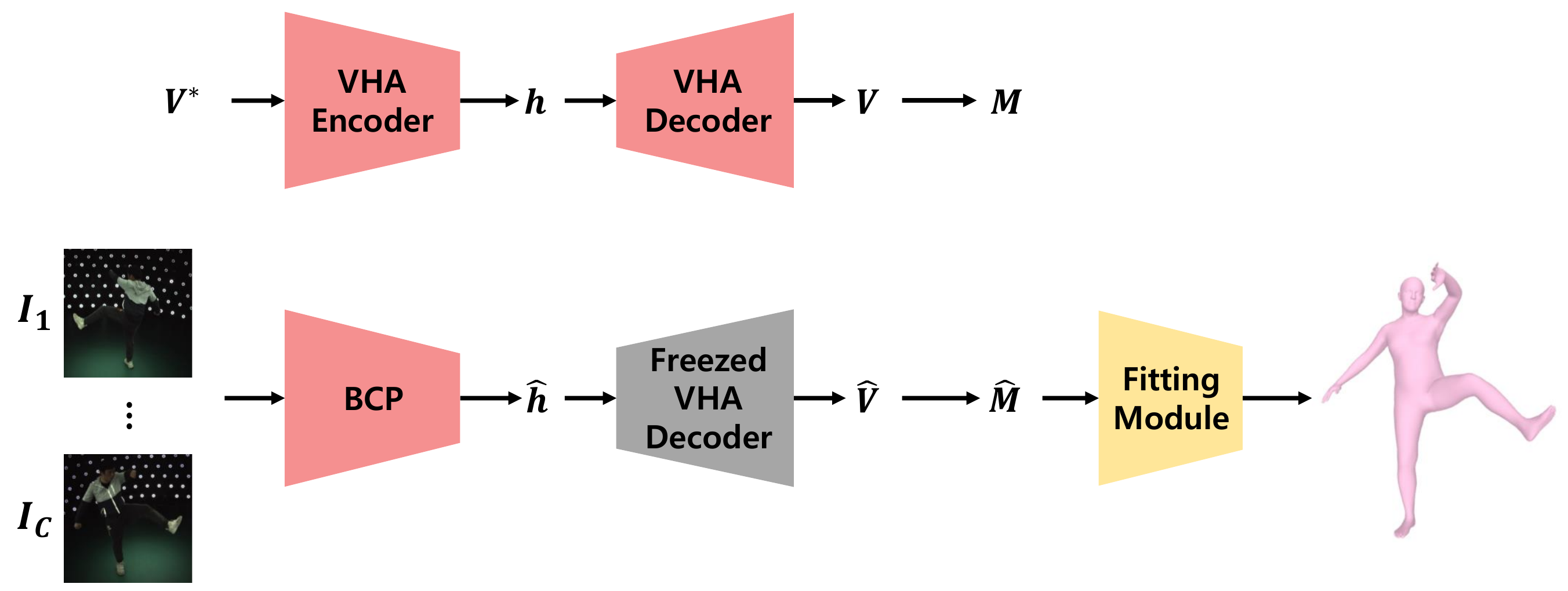}
%\vspace*{-4mm}
\caption{Overall structure of the proposed method. Modules containing learnable parameters are colored pink.}
\label{fig:overview}
%\vspace*{-4mm}
\end{figure}

We propose a method for reconstructing an SMPL-based 3D mesh~\cite{2015_SMPL} of a single person from $C$ calibrated multi-view images. Fig.~\ref{fig:overview} illustrates the overall structure of the proposed method, which is composed of the VHA, BCP, and fitting module. In VHA, the VHA encoder first converts volumetric heatmaps $V^{*}\in{}\mathbb{R}^{{N}\times{}64\times{}64\times{}64}$ for $N$ ground-truth subsampled vertices~\cite{Chun_2023_WACV} into the low-dimensional latent code $h\in{}\mathbb{R}^{256\times{}4\times{}4\times{}4}$, from which volumetric heatmaps $V\in{}\mathbb{R}^{{N}\times{}64\times{}64\times{}64}$ are reconstructed by the VHA decoder. From $V$, the 3D coordinates of subsampled mesh vertices $M\in{}\mathbb{R}^{N\times{}3}$ are obtained through the 3D soft-argmax operation~\cite{2018_Sun, 2019_LT, Chun_2023_WACV}. The latent code $h$ is used for training BCP, which predicts the latent code $\hat{h}\in{}\mathbb{R}^{256\times{}4\times{}4\times{}4}$ from the $C$ multi-view images $\{I_{c}\in{}\mathbb{R}^{H_{0}\times{}W_{0}\times{}3}\}_{c=1}^{C}$. The predicted latent code $\hat{h}$ is fed into the VHA decoder to produce the volumetric heatmaps $\hat{V}\in{}\mathbb{R}^{N\times{}64\times{}64\times{}64}$, from which the vertex coordinates $\hat{M}\in{}\mathbb{R}^{N\times{}3}$ are computed through the 3D soft-argmax operation. Finally, the SMPL model is fitted to $\hat{M}$ by the fitting module, and the SMPL parameters corresponding to the target person are obtained.

\subsection{Vertex Heatmap Representation}
\label{ssec:heatmap_representation}

%In this subsection, we describe the method for generating volumetric heatmaps $V^{*}$ from the coordinates of ground-truth vertices $M^{*}\in{}\mathbb{R}^{N\times{}3}$, where $*$ denotes the ground truth.
Volumetric heatmaps $V^{*}$ are generated from the coordinates of ground-truth vertices $M^{*}\in{}\mathbb{R}^{N\times{}3}$, where $*$ denotes the ground truth. We define a cuboid containing $M^{*}$, with an edge length of $L$, centered on the pelvis of the target person. We then voxelize the cuboid into a $D\times{}H\times{}W$ resolution, where $D$, $H$, and $W$ denote the $z$-, $y$-, and $x$-axis resolutions, respectively. The process of calculating the vertex heatmap $V_{n}^{*}\in{}\mathbb{R}^{64\times{}64\times{}64}$ from the $n$-th ground-truth vertex can be written as:
\begin{equation}
\label{eq:volumetric_heatmap}
    V_n^{*}(i,j,k)=e^{-\frac{(i-i_{n}^{*})^2+(j-j_{n}^{*})^2+(k-k_{n}^{*})^2}{2\sigma^{2}}},
\end{equation}
where $(i_{n}^{*},j_{n}^{*},k_{n}^{*})$ denotes the index of the voxel containing the $n$-th ground-truth vertex.

\begin{figure}[t]
\centering
\includegraphics[width=\linewidth]{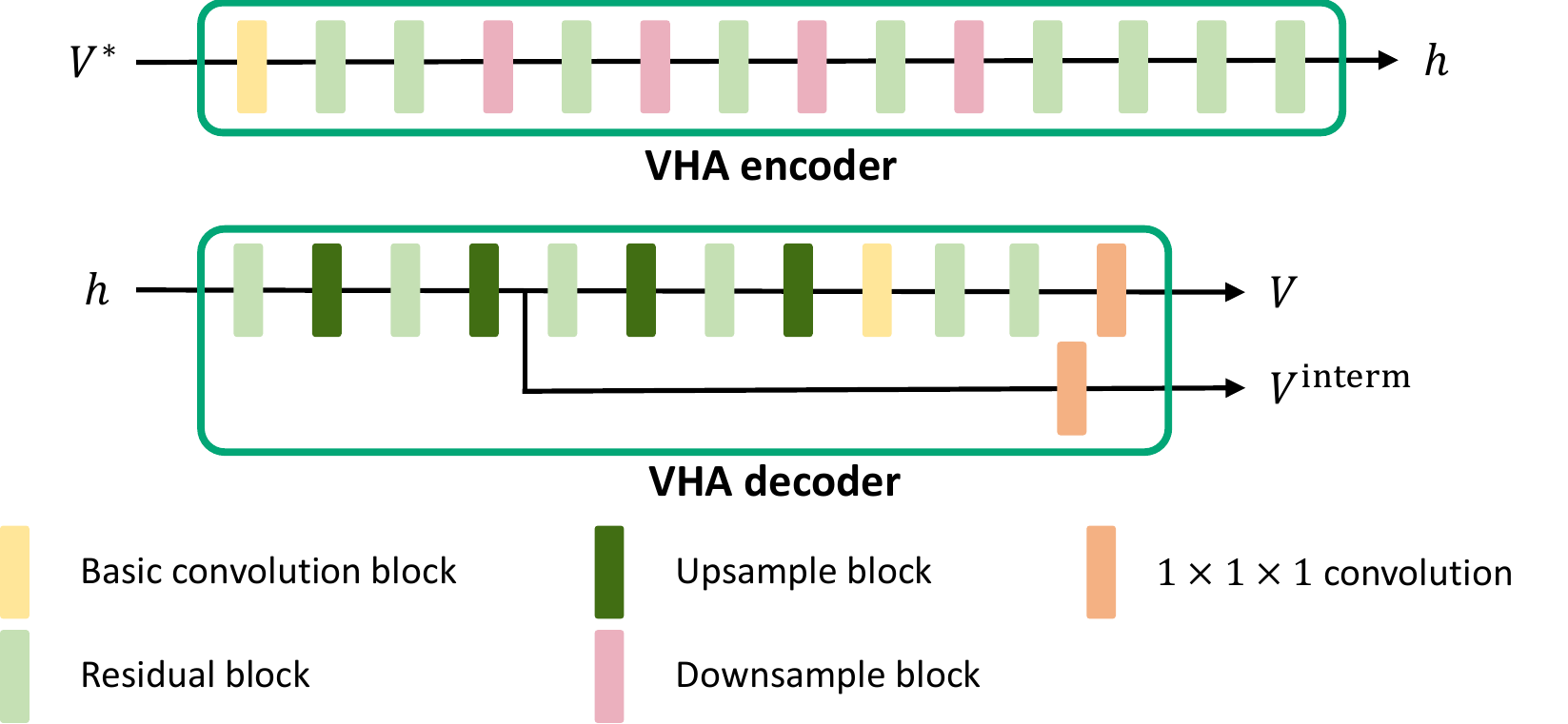}
%\vspace*{-4mm}
\caption{The architecture of the vertex heatmap autoencoder.}
\label{fig:vha}
%\vspace*{-4mm}
\end{figure}

\subsection{Vertex Heatmap Autoencoder}
\label{ssec:vha}

Fig.~\ref{fig:vha} shows the detailed structure of the VHA, which consists of a sequential combination of encoder and decoder. The VHA encoder computes the latent code $h$ from the vertex heatmap $V^{*}$. The VHA decoder reconstructs two vertex heatmaps $V\in{}\mathbb{R}^{N\times{}64\times{}64\times{}64}$ and $V^{\text{interm}}\in{}\mathbb{R}^{N\times{}16\times{}16\times{}16}$, from the latent code $h$. VHA consists of a combination of basic convolution block, residual block, downsample block, and $1\times{}1\times{}1$ convolution block, as shown in Fig.~\ref{fig:vha}. The basic convolution block, residual block, downsample block, and upsample block have the same structure as those in LMT~\cite{Chun_2023_WACV}.
%The VHA encoder consists of a basic convolution block, nine residual blocks, and four downsample blocks. The VHA decoder consists of a basic convolution block, six residual blocks, four upsample blocks, and two $1\times{}1\times{}1$ convolution blocks.

To train the VHA, $V$ and $V^{\text{interm}}$ are converted into vertex coordinates $M$ and $M^{\text{interm}}\in{}\mathbb{R}^{N\times{}3}$ through the 3D soft-argmax operation, respectively. Details of the 3D soft-argmax operation can be found in~\cite{2018_Sun, 2019_LT, Chun_2023_WACV}. We have empirically found that applying L1 loss not only to $M$ but also to the intermediate mesh $M^{\text{interm}}$ helps reduce the reconstruction error of VHA. The loss function for training VHA can be written as follows:
\begin{equation}
\label{eq:vha_loss_all}
    \mathcal{L}_{\text{VHA}}=\mathcal{L}_{M}+\mathcal{L}_{M^{\text{interm}}},
\end{equation}
\begin{equation}
\label{eq:vha_loss}
    \mathcal{L}_{M}=\frac{1}{N}\sum_{n=1}^{N}\|M_{n}-M_{n}^{*}\|_{1},
\end{equation}
\begin{equation}
\label{eq:vha_loss_intermediate}
    \mathcal{L}_{M^{\text{interm}}}=\frac{1}{N}\sum_{n=1}^{N}\|M_{n}^{\text{interm}}-M_{n}^{*}\|_{1},
\end{equation}
where $M_n$, $M^{*}_n$, and $M^{\text{interm}}_n$ denote the $n$-th row vector of $M$, $M^{*}$, and $M^{\text{interm}}$, respectively.

%The process of obtaining $M$ from $V$ can be written as:
%\begin{equation}
%\label{eq:softargmax_1}
%    \tilde{V}_{n}=\frac{\exp(V_{n})}{\sum_{i,j,k}\exp(V_{n}(i,j,k))},
%\end{equation}
%\begin{equation}
%\label{eq:softargmax_2}
%    M_{n}=\sum_{i,j,k}r\cdot{}\tilde{V}_{n}(i,j,k),
%\end{equation}
%where $r=[r_{i},r_{j},r_{k}]$ and $M_n$ denote the world coordinate vector of the voxel with index $(i,j,k)$ and the $n$-th row vector of $M$, respectively. The process of calculating $M^{interm}$ is identical to that of $M$.

\subsection{Body Code Predictor}
\label{ssec:bcp}

We borrow the design of LMT~\cite{Chun_2023_WACV} to construct BCP. However, the output of BCP is not the vertex heatmap but the latent code unlike LMT. Therefore, we replace the vertex regression module in the LMT with the 3D CNN encoder. As a result, BCP consists of the visibility module, CNN backbone, feature aggregation module, and 3D CNN encoder. The 3D CNN encoder has the same structure as the VHA encoder. Details of the rest of the modules can be found in~\cite{Chun_2023_WACV}. %BCP has the same structure as LMT except for the Vertex Regression Module [2]. Thus, it consists of the visibility module, CNN backbone, feature aggregation module, and [2]. And it includes a 3D CNN encoder, instead of a Vertex Regression Module. The 3D CNN encoder has the same structure as the VHA encoder.

From the input multi-view images $\{I_c\}_{c=1}^{C}$, the visibility module first computes per-vertex visibility vectors. The obtained visibility vectors and the input images are fed into the CNN backbone to generate $C$ 2D features. The $C$ 2D features are fed into the feature aggregation module to produce a volumetric feature. The volumetric feature is converted into the latent code $\hat{h}$ through the 3D CNN encoder. We can obtain 3D human mesh vertices $\hat{M}$ by sequentially applying the pretrained VHA decoder and 3D soft-argmax to the predicted latent code $\hat{h}$.

We apply the L1 loss to the predicted latent code $\hat{h}$ to train BCP. Also, we have empirically found that applying the additional L1 loss to the vertex coordinates $\hat{M}$ can help in training BCP. The losses for training BCP can be written as:
\begin{equation}
\label{eq:bcp_loss}
    \mathcal{L}_{\text{BCP}}=\lambda\mathcal{L}_{\hat{h}}+\mathcal{L}_{\hat{M}},
\end{equation}
\begin{equation}
\label{eq:bcp_code_loss}
    \mathcal{L}_{\hat{h}}=\frac{1}{N_{h}}\sum_{i,j,k,w}\|\hat{h}(i,j,k,w)-h(i,j,k,w)\|_{1},
\end{equation}
\begin{equation}
\label{eq:bcp_vtx_loss}
    \mathcal{L}_{\hat{M}}=\frac{1}{N}\sum_{n=1}^{N}\|\hat{M}_{n}-M_{n}^{*}\|_{1},
\end{equation}
where $(i,j,k,w)$, $N_{h}$, and $\lambda$ denote the voxel index, the number of voxels in the latent code, and the weight of the code loss, respectively. In our current implementation, $N_h=256\times{}4\times{}4\times{}4=16384$.

\subsection{Fitting Module}
\label{ssec:fitting_module}

The fitting module estimates the SMPL parameters $\beta\in{}\mathbb{R}^{10}$, $\theta\in{}\mathbb{R}^{72}$, and $t\in{}\mathbb{R}^{3}$ by fitting the SMPL model to $\hat{M}$ predicted by the BCP and VHA decoder. Here, $\beta$, $\theta$, and $t$ denote the SMPL shape parameter, SMPL pose parameter, and global translation, respectively. In the fitting module, the SMPL mesh $M^{\text{fit}}=\mathcal{M}(\beta,\theta,t)\in{}\mathbb{R}^{6890\times{}3}$ is first computed from the SMPL parameters and then converted into the subsampled vertices $M^{\text{fit}}_{\text{sub}}\in{}\mathbb{R}^{N\times{}3}$~\cite{COMA_ECCV18, Chun_2023_WACV}. The distance between $M^{\text{fit}}_{\text{sub}}$ and $\hat{M}$ is defined as the data term $\mathcal{E}_{\text{data}}$. It is used as a cost function for optimization with the regularization term $\mathcal{E}_{\text{reg}}$ that prevents unnatural poses from being generated. Details of the fitting module can be found in~\cite{Chun_2023_WACV}. The fitted mesh $M^{\text{fit}}$ obtained through optimization is used together with joint coordinates $J=GM^{\text{fit}}\in{}\mathbb{R}^{17\times{}3}$ for the final evaluation of the proposed method, where $G\in{}\mathbb{R}^{17\times{}6890}$ is the pretrained joint regression matrix.

\section{Experimental Results}
\label{sec:experimental_results}

\subsection{Implementation Details}
\label{ssec:implementation_details}

The bounding boxes provided in the datasets are used to crop the human regions from the input images. We first train the VHA to obtain the latent code $h$ required for training the BCP. The learning rate and mini-batch size for VHA training are set to 1e-3 and 10, respectively. The learning rate and mini-batch size for BCP training are set to 1e-4 and 3, respectively. No augmentation technique is used to train VHA and BCP. The 3D CNN encoder of BCP is initialized with the weights of the pretrained VHA encoder and then fine-tuned. We set the number of subsampled vertices $N$, the edge length $L$ and resolutions $D$, $H$, $W$ of the cuboid, the height $H_0$ and width $W_0$ of the input image, and the code loss weight $\lambda$ to 108, 2.0m, 64, 64, 64, 384, 384, and 100.0, respectively. We use the Adam optimizer~\cite{2015_Kingma} to train VHA and BCP for 55 and 20 epochs, which takes about 16 and 4.5 days using a single RTX 3090 GPU, respectively.

\subsection{Datasets}
\label{ssec:datasets}

We train VHA using the Human3.6M~\cite{2014_H36M} training set and AMASS~\cite{moshpp_AMASS_2019}, which are large-scale datasets. BCP is trained and evaluated on each of the Human3.6M and LightStage~\cite{peng2021neural} datasets. The AMASS sequence data are sampled every 10 frames, and the ground-truth mesh provided in the dataset is used for training. According to the Human3.6M general setting~\cite{2018_Kanazawa, 2019_LT, Chun_2023_WACV, 2020_Shin, choi2022learning}, S1, S5, S6, S7, and S8 are used as train data, and S9 and S11 are used as test data. The meshes obtained by applying MoSh~\cite{mosh_Loper_2014}, as the ground truth of Human3.6M, are used as train and test data~\cite{Chun_2023_WACV}. We use the LightStage dataset to investigate the training capability of the model on small-scale datasets. All sequences of LightStage are sampled every 5 frames and then used. About 2700 frames obtained from sequences 363-371 and 377-384 are used as train data, and about 500 frames obtained from sequences 385-388 are used as test data. We use only images obtained from cameras 1, 7, 13, and 19 among all the cameras in LightStage for training and evaluation. LightStage provides the SMPLX~\cite{SMPLify_X_2019_Pavlakos} parameters as the ground truth, not the SMPL parameters. Therefore, we use the parameter converting function~\footnote{https://github.com/vchoutas/smplx} released from~\cite{SMPLify_X_2019_Pavlakos} to convert SMPLX parameters to SMPL parameters and utilize the result as the ground truth for training and evaluation. We evaluate the joint location, joint rotation, and human mesh vertex estimation performance of the proposed method using the ground-truth mesh of each dataset through MPJPE, angular distance~\cite{Hartley_IJCV2013}, and MPVE metrics, respectively. Their units are mm, degree, and mm.
% ~\cite{2018_Kanazawa, 2019_Kolotouros_CVPR, 2019_Kolotouros_ICCV, 2019_LT, Chun_2023_WACV, 2020_I2L, 2020_Shin, 2021_Lin_CVPR, 2021_Lin_ICCV, cho2022FastMETRO, choi2022learning, Choi_2020_ECCV_Pose2Mesh}
%%%%%%%%%%%%%%%%%%%%%%%%%%%%%%%%%%%%%%%%%%%%%%%%%%%%%%%%%%%%%%%%%%%%%%
% The table for Representation Learning

\begin{table}[t]
\caption{Ablation results for representation learning on Human3.6M. The last row corresponds to our proposed model. The best results are shown in bold.}
\centering
{\footnotesize
\begin{tabular}{C{0.8cm}C{0.8cm}C{1.1cm}|C{0.85cm}C{0.8cm}C{1.0cm}}
\specialrule{.1em}{.05em}{.05em}
{$\mathcal{L}_{\hat{M}}$} & {$\mathcal{L}_{\hat{h}}$} & {Pretrain} & {MPJPE} & {MPVE} & {Angular} \\ 
\hline

{\checkmark} & {\xmark} & {\xmark} & {18.0} & {25.1} & {11.72} \\

{\checkmark} & {\checkmark} & {\xmark} & {16.7} & {24.0} & {11.29} \\

{\checkmark} & {\xmark} & {\checkmark} & {15.8} & {22.7} & \textbf{10.82} \\

{\xmark} & {\checkmark} & {\checkmark} & {16.2} & {23.1} & {10.88} \\ \hline

{\checkmark} & {\checkmark} & {\checkmark} & \textbf{15.6} & \textbf{22.3} & {10.87} \\

\specialrule{.1em}{.05em}{.05em}
\end{tabular}
}
\label{tab:representation_learning}
\end{table}

%%%%%%%%%%%%%%%%%%%%%%%%%%%%%%%%%%%%%%%%%%%%%%%%%%%%%%%%%%%%%%%%%%%%%%

%%%%%%%%%%%%%%%%%%%%%%%%%%%%%%%%%%%%%%%%%%%%%%%%%%%%%%%%%%%%%%%%%%%%%%
% Comparison with SOTA methods
%
\begin{table}[t]
\caption{Comparison with the state-of-the-art methods. Numbers in parentheses represent the results of cross-dataset generalization experiments, where training and evaluation are performed on Human3.6M and LightStage, respectively. ``$\natural$'' indicates that VHA is trained without the AMASS dataset.}
\scriptsize
\centering
\setlength\tabcolsep{1.0pt}
\def\arraystretch{1.1}
\begin{tabular}{L{2.1cm}|C{0.75cm}C{0.70cm}C{0.9cm}|C{1.2cm}C{1.2cm}C{1.35cm}}
\specialrule{.1em}{.05em}{.05em}
\multirow{2}{*}{Model} & \multicolumn{3}{c|}{Human3.6M} & \multicolumn{3}{c}{LightStage}  \\ \cline{2-7}
{} & {MPJPE} & {MPVE} & {Angular} & {MPJPE} & {MPVE} & {Angular} \\ \hline

\multicolumn{7}{l}{Backbone: ResNet50, Input Size: $224\times{}224$} \\ \hline

{Param. Regr.~\cite{2020_Shin}} & {46.9} & {-} & {-} & {-} & {-} & {-}\\

{LMT~\cite{Chun_2023_WACV}} & {30.6} & {42.3} & {14.61} & {31.8} & {40.2} & {18.56}\\

{VHA+BCP (Ours)} & \bf{26.8} & \bf{39.6} & \bf{13.88} & \bf{18.2} & \bf{23.4} & \bf{16.48}\\ \hline \hline

\multicolumn{7}{l}{Backbone: ResNet152, Input Size: $384\times{}384$} \\ \hline

{LT-fitting~\cite{2019_LT,lightcap2021}} & {16.2} & {35.2} & {15.73} & {13.8 (40.9)} & {31.5 (60.1)} & {20.00 (24.49)} \\

{LMT~\cite{Chun_2023_WACV}} & {17.6} & {23.7} & {11.33} & {19.0 (33.0)} & {26.0 (40.6)} & {16.70 (18.54)} \\

{VHA+BCP (Ours) $\natural$} & {15.7} & {22.7} & \bf{10.83} & {13.9 (34.1)} & {19.3 (42.7)} & \textbf{16.08} (19.79) \\

{VHA+BCP (Ours)} & \bf{15.6} & \bf{22.3} & {10.87} & \textbf{13.2} (\textbf{29.9}) & \textbf{18.6} (\textbf{37.7}) & {16.15} (\textbf{18.53}) \\

\specialrule{.1em}{.05em}{.05em}
\end{tabular}
\label{tab:comparison_sota}
\end{table}

%%%%%%%%%%%%%%%%%%%%%%%%%%%%%%%%%%%%%%%%%%%%%%%%%%%%%%%%%%%%%%%%%%%%%%

\subsection{Ablation Experiments}
\label{ssec:ablation_experiments}

The human body prior learned through VHA is utilized for mesh reconstruction by BCP in two ways: supervision of BCP using latent code and initialization of BCP weights with pretrained autoencoder weights. We perform ablation experiments to verify the effect of such a human body prior. The performances of BCPs learned and evaluated on Human3.6M train and test data are shown in Table~\ref{tab:representation_learning}. The first, second, and third columns of the table denote whether the vertex loss $\mathcal{L}_{\hat{M}}$, code loss $\mathcal{L}_{\hat{h}}$, and pretrained autoencoder weights are used for BCP training, respectively. According to the experimental results, the additional use of the proposed body prior improves the BCP performance compared with BCP trained by applying only vertex loss. As a result, the combination of vertex-level supervision and body prior shows the best mesh reconstruction performance.

\subsection{Comparison with State-of-the-art Methods}
\label{ssec:comparison_sota}

Table~\ref{tab:comparison_sota} compares the proposed model with existing methods for estimating SMPL parameters from calibrated multi-view images on Human3.6M and LightStage datasets. The numbers in parentheses show the evaluation results of BCP and existing models for cross-dataset generalization performance. In this experiment, models are trained on Human3.6M train data and evaluated on all train and test data of LightStage. The remaining numbers in Table~\ref{tab:comparison_sota} denote the results of the models trained and tested on each dataset.

According to the results from Human3.6M, the proposed method outperforms the existing hybrid (i.e., deep learning $+$ optimization)~\cite{Chun_2023_WACV, 2019_LT, lightcap2021} and parameter regression~\cite{2020_Shin} approaches. This result demonstrates that using the human body prior helps train the model effectively on a large-scale dataset such as Human3.6M.

We use the small-scale LightStage dataset for training and testing to demonstrate that the proposed body prior enables the model to learn efficiently even from a limited amount of train data. The key difference between LMT and the proposed method lies in using prior knowledge in the learning-based vertex regression stage, which significantly improves the joint and vertex location estimation accuracy of the proposed model, as shown in Table~\ref{tab:comparison_sota}. Unlike the proposed method, LT-fitting estimates human joints, not human surfaces, in the learning-based regression stage. This joint location does not provide sufficient information to resolve the ambiguity of joint rotation and human shape reconstruction~\cite{Chun_2023_WACV}. As a result, the proposed method significantly outperforms LT-fitting in terms of MPVE and angular distance.

The LightStage dataset includes human poses that are excluded in Human3.6M and subjects wearing clothes that do not reveal their body shape. Therefore, achieving good reconstruction performance of the model trained in Human3.6M on LightStage requires a high generalization capability. The numbers in the parentheses in Table~\ref{tab:comparison_sota} show that using the human body prior helps model inference under this cross-dataset setting. Using a large-scale motion capture dataset including various human poses and body shapes, such as AMASS, also helps in representation learning and further improves the generalization performance of the model, which is shown in the last two rows of Table~\ref{tab:comparison_sota}.

\section{Conclusion}
\label{sec:conclusion}

This study proposes autoencoder-based representation learning of volumetric heatmaps for 3D human mesh reconstruction. The proposed model (i.e., VHA) serves as an effective body prior to improve the cross-dataset generalization capability and training data efficiency of the mesh reconstruction method (i.e., BCP), achieving state-of-the-art performance. Future work will investigate whether the proposed body prior is effective for mesh reconstruction from in-the-wild images.

%Recently, in the task of SMPL-based human mesh reconstruction from multi-view images, the hybrid approach that first estimates human geometry using a volumetric heatmap-based method and then combines it with an optimization-based method has shown good performance. However, such methods tend to have poor performance when they receive unseen test poses as input or when they are not trained through enough data. This is because using human body prior only in the optimization process has limitations in refining the wrongly estimated human geometry from learning-based methods. In order to solve this problem, in this paper, a method of using body prior in learning-based method through autoencoder-based representation learning for volumetric heatmap is proposed. The proposed method has been proven to be effective in utilizing body prior during training, as it outperforms previous methods for human pose and mesh reconstruction from multi-view images on the Human3.6M and LightStage datasets. However, the proposed method is limited to the volumetric heatmap approach and further exploration is needed to utilize prior knowledge in more diverse methods. Additionally, exploring methods for human pose and mesh reconstruction from in-the-wild input images remains a topic of ongoing research as a future work.

\bibliographystyle{IEEEbib}
\bibliography{strings,refs}

\end{document}